\documentclass{article}

\newif\ifcomments

\commentsfalse
\ifcomments
\usepackage[draft]{hyperref}
\newcommand{\comments}[1]{#1}
\else
\usepackage{hyperref}
\newcommand{\comments}[1]{}
\fi
\usepackage{array}
\usepackage{times}
\usepackage{graphicx}
\usepackage{subcaption} 
\usepackage{natbib}
\usepackage{amsmath}
\usepackage{amssymb}
\usepackage{xcolor}
\usepackage{float}
\usepackage{ownstyles}
\usepackage{appendix}

\usepackage[accepted]{icml2017}

\newcommand{\imageX}{\mathbf{x}}
\newcommand{\imageSetX}{\mathbf{X}}
\newcommand{\patchY}{\mathbf{y}}
\newcommand{\patchSetY}[1]{{\mathbf{Y}_{#1 \times #1}}}
\newcommand{\latentZ}{\mathbf{z}}
\newcommand{\patchG}{\mathbf{g}}
\newcommand{\gridG}[1]{{\mathbf{G}_{#1 \times #1}}}

\newcommand{\B}{\bfseries}
\newcommand{\tast}{\makebox[0pt][l]{\textsuperscript{*}}}
\newcolumntype{M}{>{\centering\arraybackslash}m}
\newcounter{rowcounter}[table]
\newcommand{\nextrow}[1]{{\refstepcounter{rowcounter}\label{#1}}}

\icmltitlerunning{Spatial PixelCNN}

\begin{document} 

\twocolumn[
\icmltitle{Spatial PixelCNN: Generating Images from Patches}



% you can specify symbols, otherwise they are numbered in order

\begin{icmlauthorlist}
  \icmlauthor{Nader Akoury}{}{dojoteef@gmail.com}
  \icmlauthor{Anh Nguyen}{au}{anhnguyen@auburn.edu}
\end{icmlauthorlist}

\icmlaffiliation{au}{Auburn University, USA}

\icmlcorrespondingauthor{Nader Akoury}{dojoteef@gmail.com}

\icmlkeywords{PixelCNN, VAE, machine learning, ICML}

\vskip 0.3in
]



\printAffiliationsAndNotice{}  % leave blank if no need to mention equal contribution

\begin{abstract} 
  In this paper we propose Spatial PixelCNN, a conditional autoregressive model that generates images from small patches.
  By conditioning on a grid of pixel coordinates and global features extracted from a Variational Autoencoder (VAE), we are able to train on patches of images, and reproduce the full-sized image.
  We show that it not only allows for generating high quality samples at the same resolution as the underlying dataset, but is also capable of upscaling images to arbitrary resolutions (tested at resolutions up to $50\times$) on the MNIST dataset.
  Compared to a PixelCNN++ baseline, Spatial PixelCNN quantitatively and qualitatively achieves similar performance on the MNIST dataset.
\end{abstract} 

\section{Introduction}

% Put this here so it appears on the first page
\begin{figure}[tb]
  \centering
  \begin{subfigure}[b]{1\linewidth}
    \centering
    \includegraphics[width=1.0\linewidth]{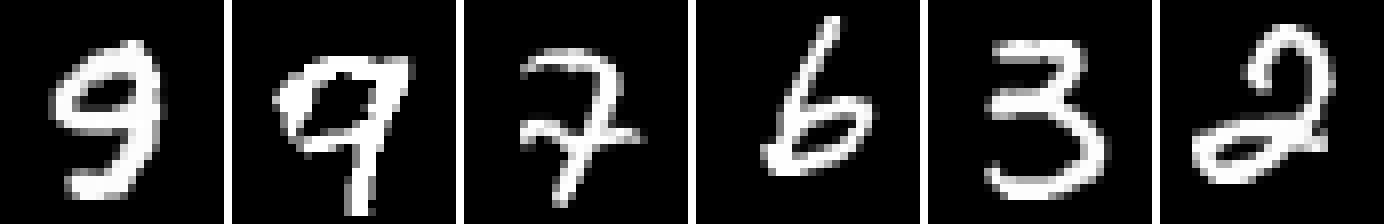}
    \caption{PixelCNN++ \cite{salimans2017pixelcnn++}}
    \label{fig:random-samples-pixelcnn}
    \vspace{0.1cm}
  \end{subfigure}
  \begin{subfigure}[b]{1\linewidth}
    \centering
    \includegraphics[width=1.0\linewidth]{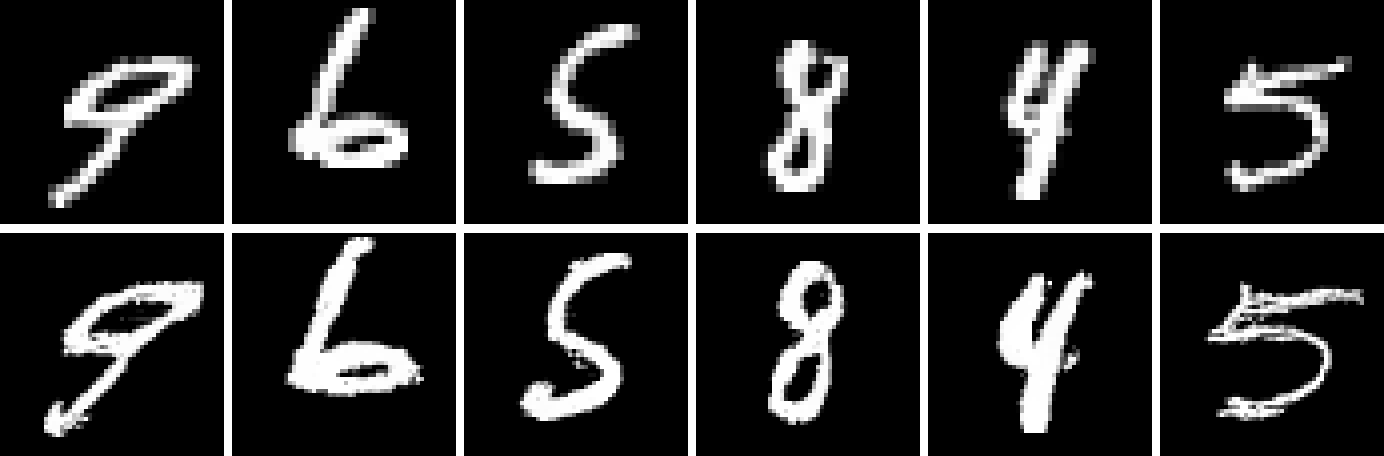}
    \caption{Spatial PixelCNN (small network) trained on $8 \times 8$ patches.}
    \label{fig:random-samples-small}
    \vspace{0.1cm}
  \end{subfigure}
  \begin{subfigure}[b]{1\linewidth}
    \centering
    \includegraphics[width=1.0\linewidth]{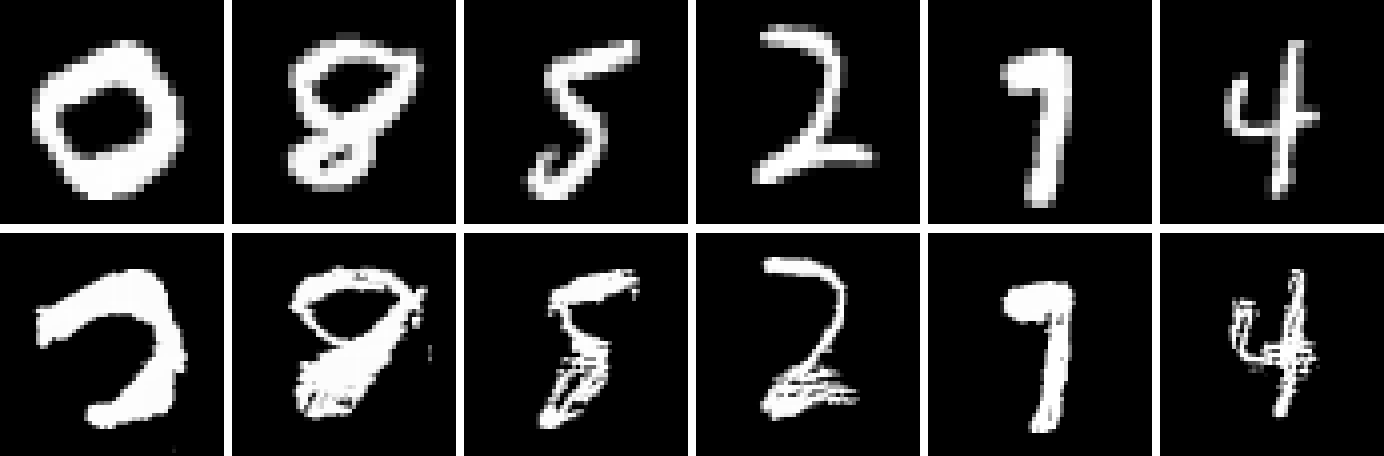}
    \caption{Spatial PixelCNN (large network) trained on $16 \times 16$ patches.}
    \label{fig:random-samples-large}
  \end{subfigure}
  \caption{
    Random samples generated from PixelCNN++ and Spatial PixelCNN.
    Samples on the top row of each group are generated at $28 \times 28$ resolution, while the bottom are $56 \times 56$ resolution.
    All three models generate high quality samples at $28 \times 28$ resolution.
    Notably, Spatial PixelCNN qualitatively performs on par with PixelCNN++, despite training on $8 \times 8$ patches.
    Compared to training on $16 \times 16$ patches, Spatial PixelCNN trained on $8 \times 8$ patches produces more coherent $56 \times 56$ resolution samples.
  }
  \label{fig:random-samples}
\end{figure}

% Put this here so it appears on the second page
\begin{figure*}[ht]
  \centering
  \includegraphics[width=1.0\linewidth]{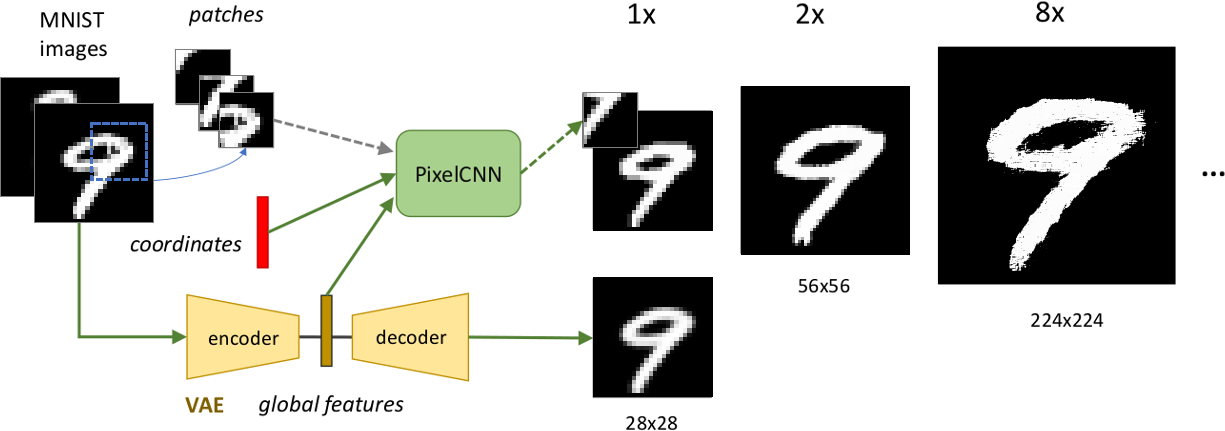}
  \caption{
    An overview of our model.
    We train a VAE on MNIST digits (upper left) to extract global features (lower left), and reconstruct its input (lower right).
    Our Spatial PixelCNN then models the images using patches (upper left), conditioned on coordinates and the learned global features.
    We show reconstructions\footnotemark (upper right) at $28 \times 28$, $56 \times 56$, and $224 \times 224$ resolution (an $8\times$ upscaling factor).
    More reconstructions are in Fig.~\ref{fig:reconstructions} and Appendix~\ref{app:large-upscaling}.
  }
  \label{fig:concept}
\end{figure*}

Generative image modeling has elicited much excitement in the past few years.
Much of the enthusiasm is centered on a small set of popular techniques.
These include variational inference, through the use of the reparameterization trick \cite{Kingma:2013aa,pmlr-v32-rezende14}, integral probability metrics \cite{goodfellow2014generative,Arjovsky:2017aa,nowozin2016f}, and autoregressive, explicit density estimation \cite{oord2016pixel}.
Despite the success of these techniques, it is still challenging to generate high-resolution images, especially for datasets with large variability \cite{nguyen2017plug,odena2016conditional}, though that is readily changing \cite{karras2017progressive}.

In this paper, we propose Spatial PixelCNN, a conditional autoregressive PixelCNN \cite{oord2016pixel} capable of generating large images from small patches.
We combine the strengths of three components: (1) an autoregressive model---specifically, PixelCNN++ \cite{salimans2017pixelcnn++}---to capture local image statistics from patches; (2) a latent variable model---in our case a variational autoencoder \cite{Kingma:2013aa}---to capture the global structures in images; and (3) spatial locations of each pixel in an image.
Intuitively, an image is modeled autogressively pixel-by-pixel.
Each pixel sampled is conditioned on (1) a subset of previously sampled pixels in its neighborhood; (2) a latent code that encodes global image statistics; and (3) 2-D spatial coordinates indicating its location in the image.
This spatial conditioning enables us to reduce the coupling of each pixel from the resolution of the image.
At generation time, our approach takes a target resolution as input, and outputs an image of arbitrary size (e.g. $224\times224$ images from $8\times8$ training patches, Fig.~\ref{fig:concept}).

While performing impressively, state-of-the-art super-resolution techniques \cite{ledig2016photo,tyka2017semantic} (a) require a large dataset of high-resolution images---which may not be always available in practice; (b) have a fixed-sized output.
We instead explore learning a generic upscaling function from only low-resolution images, in the absence of high-resolution images.
Compared to existing image generative models that output fixed-sized images, our approach can be trained on small patches as opposed to full-sized images, thus requiring much less GPU memory---an important implication for scalability. 
Our method also enables the possibility of training on a dataset of images of mixed resolutions.

Our contributions are summarized as follows:

\begin{enumerate}
	\item We show the effects of spatial conditioning, using pixel coordinates, on generating high-resolution images.

	\item We show quantitative and qualitative evidence that Spatial PixelCNN models full-sized $28\times28$ MNIST \cite{lecun1998gradient} images relatively well compared to a baseline PixelCNN++ \cite{salimans2017pixelcnn++} model, despite being trained on $8\times8$ patches.

  \item We show that remarkably, Spatial PixelCNN is capable of generating coherent images at arbitrary resolutions for the MNIST dataset.

\end{enumerate}

\section{Related Work}

\footnotetext{Note that while not explicitly forced to reconstruct the VAE inputs, Spatial PixelCNN samples are very similar to VAE reconstructions when conditioned on the same latent code.}

% Put this here so it appears on the third page
\begin{figure}[b]
  \centering
  \begin{subfigure}[b]{0.2\linewidth}
    \centering
    \includegraphics[width=1.0\linewidth]{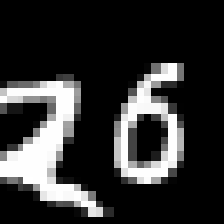}
    \caption{}
    \label{fig:image-defects-a}
  \end{subfigure}
  \hspace{0.05cm}
  \begin{subfigure}[b]{0.2\linewidth}
    \centering
    \includegraphics[width=1.0\linewidth]{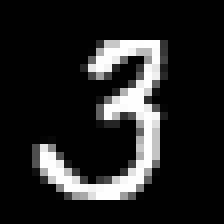}
    \caption{}
    \label{fig:image-defects-b}
  \end{subfigure}
  \hspace{0.05cm}
  \begin{subfigure}[b]{0.2\linewidth}
    \centering
    \includegraphics[width=1.0\linewidth]{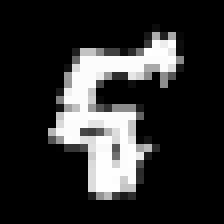}
    \caption{}
    \label{fig:image-defects-c}
  \end{subfigure}
  \hspace{0.05cm}
  \begin{subfigure}[b]{0.2\linewidth}
    \centering
    \includegraphics[width=1.0\linewidth]{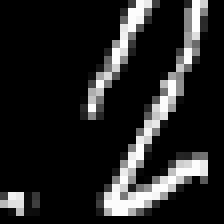}
    \caption{}
    \label{fig:image-defects-d}
  \end{subfigure}
  \caption{
    Examples showing the need for coordinates and a global latent code:
    (\textit{a}) vanilla PixelCNN++ trained on $20 \times 20$ patches, displaying a juxtaposition of two distinct digits;
    (\textit{b}) PixelCNN++ with spatial conditioning trained on $20 \times 20$ patches achieves $1.48$ bits per dimension and, models $28 \times 28$ MNIST;
    (\textit{c}) when trained on $4 \times 4$ patches, spatial conditioning alone does not confer global coherence, despite achieving $1.42$ bits per dimension;
    (\textit{d}) and while conditioning on a latent code helps with global coherence, without coordinates it does not provide scale. 
  }
  \label{fig:image-defects}
\end{figure}

Pixel Recurrent Neural Networks \cite{oord2016pixel} (PixelRNN), are a class of powerful generative model.
PixelRNN is an explicit generative model which can be trained to directly maximize the likelihood of the training data. 
Here, the likelihood $p(\imageX)$ of each 2-D image $\imageX \in \mathbb{R}^{W\times H}$ is decomposed via chain rule into:

\begin{equation}
  p(\imageX) = \prod_{j}^{n}\prod_{i}^{n}{p(x_{j \ast n+i}|\imageX_{<j \ast n+i})}
  \label{eq:autoregressive-pixelcnn}
\end{equation}

which is the product of every pixel $x_{j \ast n+i}$, conditioned on all previous pixels $\imageX_{<j \ast n+i}$ in the row-major order---a left to right, top to bottom order of columns and rows $(i, j)$.

It is this grounding in RNNs, that helps explain the motivation to train on only a patch of an image at time.
RNNs have been used extensively for sequence modeling \cite{Krause:2016aa, theis2015generative, kim2016character}.
Often in the area of natural language processing (NLP), the corpus of text being trained is too large to feed the entire sequence of characters or words to the model at once.
In order to circumvent that limitation, training involves a truncated form of backpropagation through time \cite{werbos1990backpropagation,graves2013generating}.
Our approach of training on patches can be seen as a similar trade-off.
Rather than conditioning each pixel on all previous pixels in a given image, we only condition on a local window of pixels.

Pixel Convolutional Neural Networks (PixelCNN) \cite{oord2017neural}, are a reformulation of PixelRNNs, that exploit masked convolutions to parallelize the the autoregressive computation.
The approach of using masked convolutions for autoregressive density estimation has also been exploited for text \cite{kalchbrenner2016neural}, audio \cite{oord2016wavenet}, and videos \cite{pmlr-v70-kalchbrenner17a}.

Thus, there is reason to believe sequential modeling of pixels in an image can be achieved through training on patches utilizing a PixelCNN.
Though to allow recreating the original image, there are several additional factors which are crucial to producing an adequate result.

\subsection{Variational Autoencoders}

Variational autoencoders \cite{Kingma:2013aa} are a form of latent variable model.
They are trained to encode their input $\imageX$ into a latent code $\latentZ$.
The true posterior $p(\latentZ|\imageX)$ is often intractable, so an approximate posterior $q(\latentZ|\imageX)$ is computed by optimizing a variational lower bound on the log-likelihood of the data.
Often this latent code is interpretable and encodes a global representation of $\imageX$.

\cite{gulrajani2016pixelvae,chen2016variational} have evaluated integrating a PixelCNN decoder into a VAE framework.
\citealp{gulrajani2016pixelvae} note that it is critical to ensure the receptive field of the PixelCNN is smaller than the input image $\imageX$.
Otherwise, a powerful PixelCNN decoder alone can model the entire image distribution, rendering the latent code $\latentZ$ from the VAE unused.
When evaluated on binarized MNIST \cite{larochelle2011neural}, both report the successful combination yields state-of-the-art results.

Spatial PixelCNN indeed fits the criteria needed to ensure utilization of the latent code $\latentZ$, as it only trains on a patch of an image at a time.
In fact the inclusion of a latent variable model, in this case a VAE, is necessary to ensure the model has a global representation of the images.

\subsection{Conditioning on Pixel Coordinates}

Compositional Pattern Producing Networks (CPPNs) \cite{stanley2007compositional} are feedforward networks that utilize a wide array of transfer functions and are often trained via evolutionary algorithms.
CPPNs can encode 2-D images \cite{nguyen2015deep}, 3-D objects \cite{clune2011evolving} and even weights of another target network \cite{stanley2009hypercube}.

To encode a 2-D color image, a CPPN performs a transformation $f:\mathbb{R}^{2} \to \mathbb{R}^3$ parameterized by a network which takes as input a pair of coordinates $(i, j)$ and outputs 3 color values (e.g. RGB) \cite{secretan2008picbreeder}.
A pair of coordinates $(i, j)$ is often computed by evenly sampling a pre-defined range e.g. $[-1, 1]$.
The coordinates are assembled into a grid, each corresponding to a pixel in the training image.
At image generation time, an image can be rendered at an arbitrarily large resolution by simply sampling more coordinates within the specified range, and querying the CPPN for the associated color value.

CPPNs have been shown to impose a strong spatial prior yielding images of highly regular patterns \cite{secretan2008picbreeder,nguyen2015deep}.
It is this spatial regularity from CPPNs that we exploit in this work.
We find that conditioning Spatial PixelCNN on a grid of coordinates is crucial for ensuring a coherent image is generated, and also confers an ability to upscale images.

\subsection{Other High-Resolution Image Generation Methods}

Since directly modeling high resolution images is challenging, an effective approach is to generate an image in hierarchical stages of increasing resolutions \cite{zhang2016stackgan,Denton:2015aa,karras2017progressive}.
While producing impressive results, this approach outputs fixed-sized images, and requires storing the entire image in GPU memory at once---this is problematic when the training images exceed GPU memory capacity.
Spatial PixelCNN instead allows for choosing the target image size at generation time, and only processes a small patch of the image at a time.

To cut down on GPU memory requirements, \citealp{tyka2017semantic} devises upscaling image tiles, and stitching them together to produce the final result.
While requiring less memory, the approach still retains the reliance on generating a fixed-sized output.

Our framework most closely resembles \citealp{ha2017latent}, which combines spatial coordinates, and the latent code from a VAE, trained as a GAN end-to-end on full-sized images.
Our model differs in two important ways: (1) Spatial PixelCNN conditions each pixel on the latent code \textit{and} a local neighborhood of preceding pixels rather than assuming conditional independence given the latent code; (2) Spatial PixelCNN is trained on patches rather than full-sized images.

\section{Methods}

In this section we define the mathematical formulation of our model.
We additionally describe how we generate images of a target size from the trained model.

\subsection{Spatial PixelCNN}

Our proposed model is based on a modified version of PixelCNN++ \cite{salimans2017pixelcnn++}.
In order to reduce cumbersome mathematical notation, we restate Eq.~\ref{eq:autoregressive-pixelcnn} as:

\begin{equation}
  p(\imageX) = \prod_{i}^{n^2}{p(x_{i}|\imageX_{<i})}
  \label{eq:autoregressive-pixelcnn-reformulated}
\end{equation}

where $i$ denotes the virtual index of the pixel, if the image were flattened into a 1-D array.
Rather than training the model on full-sized $n \times n$ images $\imageX$, we instead choose random patches $\patchY$ of size $m \times m$ taken from the images:

\begin{equation}
  p(\patchY) = \prod_{i}^{m^2}{p(y_{i}|\patchY_{<i})}
  \label{eq:autoregressive-patch}
\end{equation}

Given that our goal is to model the set of images $\imageSetX$, rather than merely a collection of all patches $\patchSetY{m}$, we condition on a normalized coordinate grid $\gridG{n}$ $\in \mathbb{R}^{2\times n \times n}$ that has the same number of coordinate pairs as pixels in the full-sized image.
Specifically, $\gridG{n}$ is composed of $n \times n$ evenly spaced coordinate pairs $(i, j)$ within the range $[-1, 1]$.
We choose patches $\patchG$ from $\gridG{n}$ corresponding to the given patch $\patchY$.
To condition each pixel $y_i$ on its corresponding coordinate pair $g_i$, we make use of the gated spatial conditioning as outlined in \citealp{van2016conditional} arriving at the conditional probability:

\begin{equation}
  p(\patchY|\patchG) = \prod_{i}^{m^2}{p(y_{i}|\patchY_{<i}, g_{i})}
  \label{eq:grid-conditioned-patch}
\end{equation}

We found such spatial conditioning crucial to imparting coherence to the patches $\patchY$.
Without conditioning on $\patchG$, the model may assign a high probability to distinct patches of the generated image, allowing juxtapositions (Fig.~\ref{fig:image-defects}).
However, as the model is trained on patches $\patchY$ of fewer dimensions, the extra spatial information provided by $\patchG$ is not enough to ensure global coherence (Fig.~\ref{fig:image-defects-c}).

In order to provide this global coherence, we also condition on a latent code, provided by a VAE.
Previous treatments \cite{gulrajani2016pixelvae,chen2016variational} combinining PixelCNN and VAE have made use of PixelCNN as a decoder.
We found that by only training on patches, the VAE was unable to capture global features (data not shown).
Instead, we jointly train the two models: VAE on images $\imageSetX$ and Spatial PixelCNN on patches $\patchSetY{m}$. 

Specifically, we minimize the sum of an \emph{image loss} and a \emph{patch loss}:

\begin{equation}
  \mathcal{L} = \mathcal{L}^x + \mathcal{L}^y
  \label{eq:total-loss}
\end{equation}

The \emph{image loss} $\mathcal{L}^x$ is the typical VAE loss, that of the negative log-likelihood of the images and the KL-divergence of the approximate posterior from the prior:

\begin{equation}
  \mathcal{L}^x = -\mathbb{E}_{\imageX \sim \imageSetX,\latentZ \sim q(\latentZ|\imageX)}\mathrm{log}p(\imageX) + D_{KL}(q(\latentZ|\imageX)||p(\latentZ))
  \label{eq:image-loss}
\end{equation}

The \emph{patch loss} $\mathcal{L}^y$ is the negative log-likelihood of the patches, given the coordinate grid $\patchG$ and latent $\latentZ$:

\begin{equation}
  \mathcal{L}^y = -\mathbb{E}_{\patchY \sim \patchSetY{m},\latentZ \sim q(\latentZ|\imageX)}\mathrm{log}p(\patchY|\patchG,\latentZ)
  \label{eq:patch-loss}
\end{equation}

Note that instead of co-training both VAE and Spatial PixelCNN, it is also possible to pre-train the VAE first, and then train the Spatial PixelCNN.
This pre-trained VAE typically has a lower KL-divergence than the co-trained approach.
Global features extracted from the pre-trained VAE also allow for successful reconstructions, though we do not evaluate its efficacy in this paper.

\subsection{Generating Images}

% Put this here so it appears on the fourth page
\begin{figure*}[ht]
  \centering
  \begin{subfigure}[b]{0.2\linewidth}
		\centering
    \includegraphics[width=1.0\linewidth]{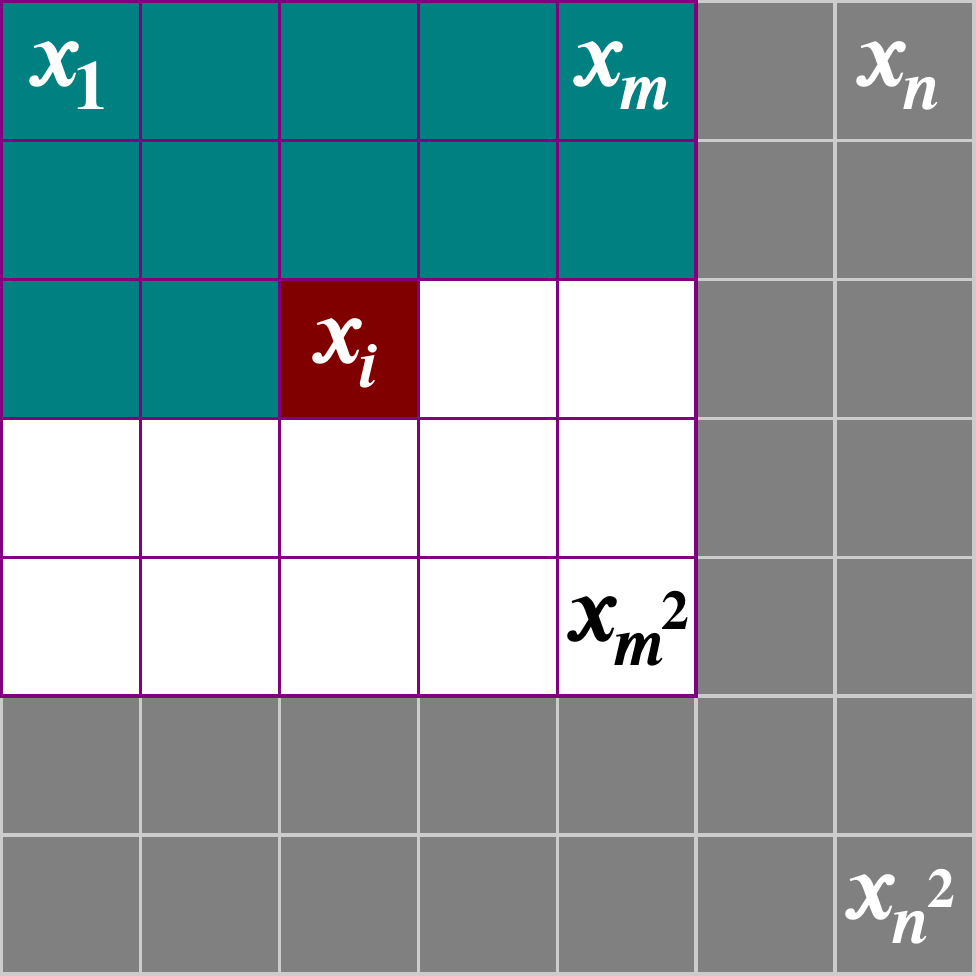}
    \caption{Maximally conditioned}
  \end{subfigure}
  \hspace{0.3cm}
  \begin{subfigure}[b]{0.2\linewidth}
		\centering
    \includegraphics[width=1.0\linewidth]{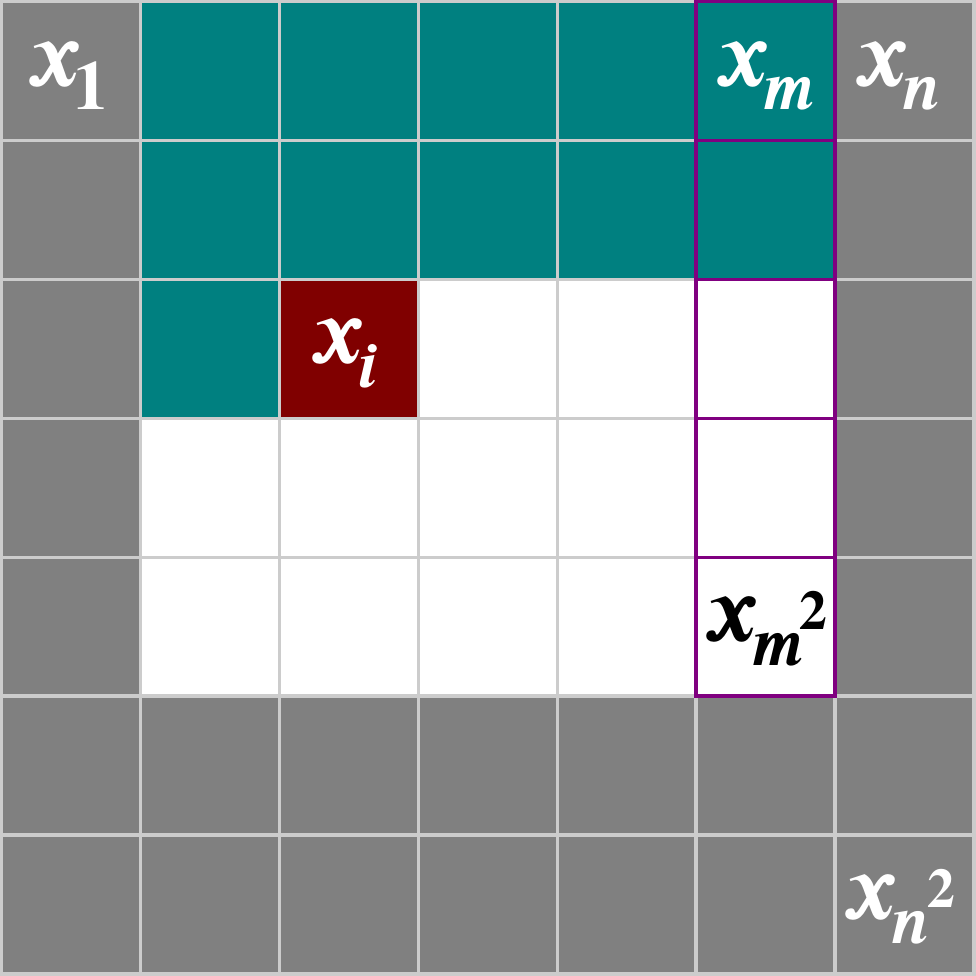}
    \caption{Column conditioned}
  \end{subfigure}
  \hspace{0.3cm}
  \begin{subfigure}[b]{0.2\linewidth}
		\centering
    \includegraphics[width=1.0\linewidth]{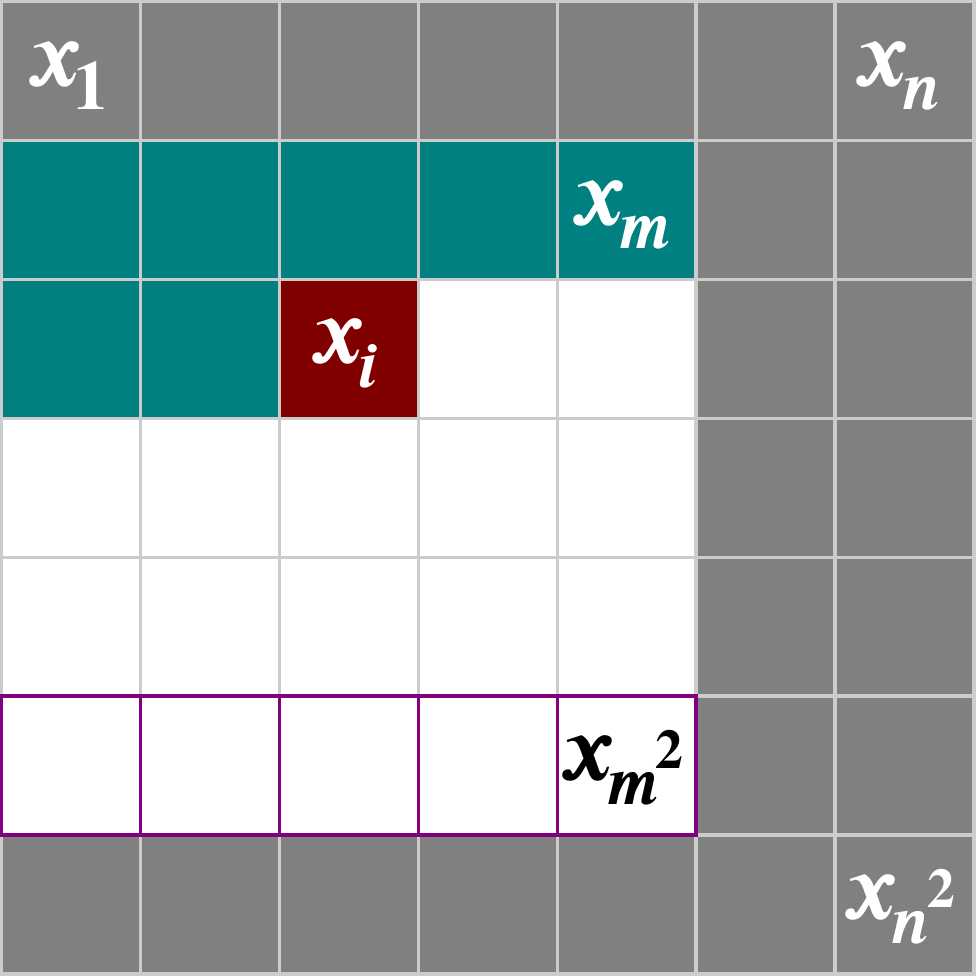}
    \caption{Row conditioned}
  \end{subfigure}
  \hspace{0.3cm}
  \begin{subfigure}[b]{0.2\linewidth}
		\centering
    \includegraphics[width=1.0\linewidth]{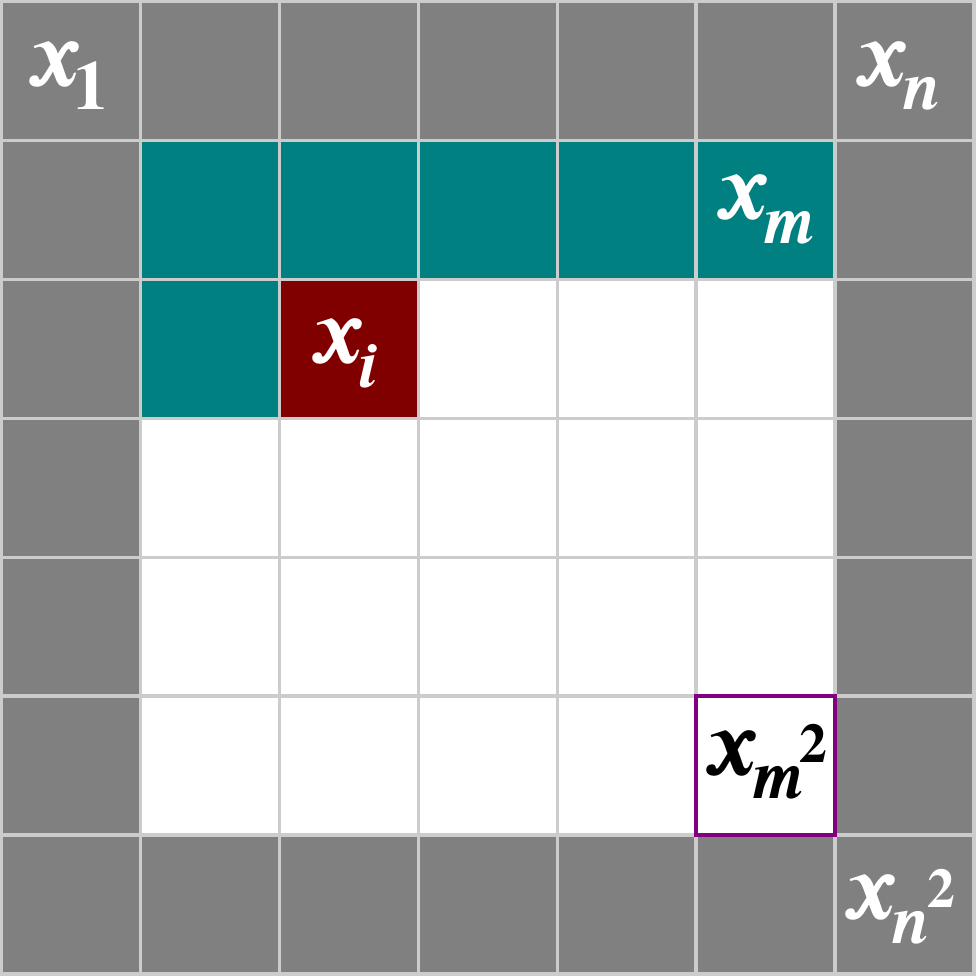}
    \caption{Pixel conditioned}
  \end{subfigure}
  \caption{
    Possible patch configurations during training and image generation.
    Grey pixels are outside the patch.
    Red pixel $x_{i}$ denotes the currently conditioned pixel during training.
    Teal pixels condition $x_{i}$.
    Purple-bordered pixels are maximally conditioned during image generation for the given patch.
  }
  \label{fig:training-patches}
\end{figure*}

As our model is based on patches of images, a sliding window is used during image generation.
In order to ensure that each pixel being generated has the maximal number of preceding pixels to condition on, the sliding window moves one pixel at a time from left to right, top to bottom (Fig.~\ref{fig:training-patches}).
This is the same ordering defined by the model for pixels to condition on (Eq.~\ref{eq:autoregressive-pixelcnn}).
Only the maximally conditioned pixels are generated from each patch of the sliding window.

One of the unique aspects of our model is its ability to upscale images to a higher resolution, while only being trained on lower resolution images.
To accomplish this, we condition on coordinate patches from a larger coordinate grid during the generation process.
That is, when we generate a $56 \times 56$ image, we subdivide the grid into $56 \times 56$ evenly spaced steps.
Then as we slide the window over the image we wish to generate, we condition on the associated patch from $\gridG{56}$.

\section{Experiments}

% Put this here so it appears with the results
\begin{figure}[ht]
  \centering
  \begin{subfigure}[b]{1\linewidth}
    \centering
    \includegraphics[width=1.0\linewidth]{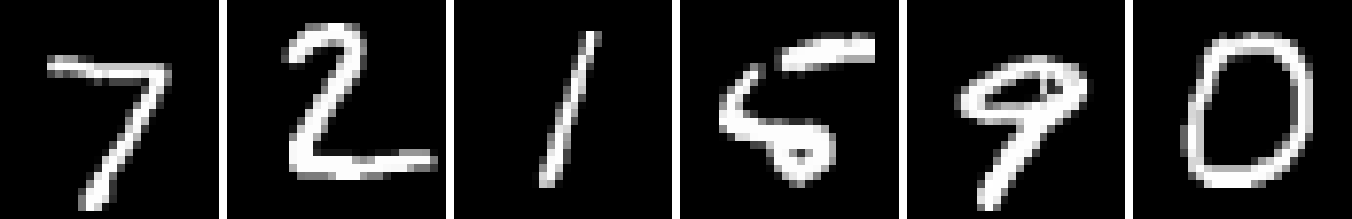}
    \caption{MNIST test images}
    \label{fig:reconstructions-original}
    \vspace{0.1cm}
  \end{subfigure}
  \begin{subfigure}[b]{1\linewidth}
    \centering
    \includegraphics[width=1.0\linewidth]{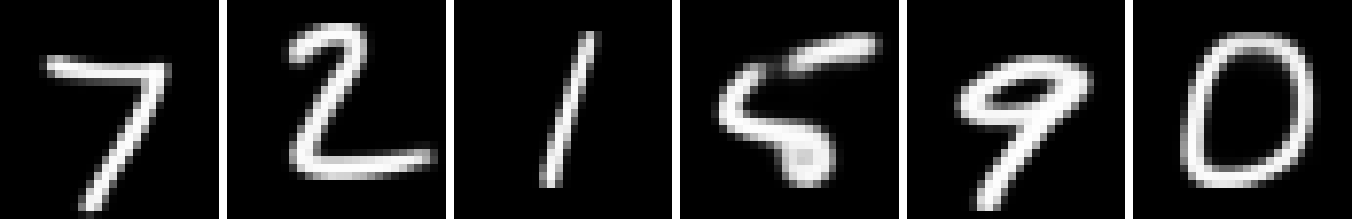}
    \caption{VAE reconstructions at $28 \times 28$}
    \label{fig:reconstructions-latent}
    \vspace{0.1cm}
  \end{subfigure}
  \begin{subfigure}[b]{1\linewidth}
    \centering
    \includegraphics[width=1.0\linewidth]{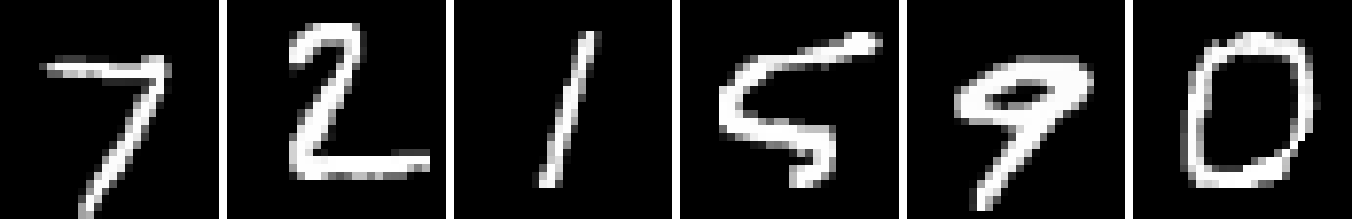}
    \caption{Reconstructions at $28 \times 28$}
    \label{fig:reconstructions-28x28}
    \vspace{0.1cm}
  \end{subfigure}
  \begin{subfigure}[b]{1\linewidth}
    \centering
    \includegraphics[width=1.0\linewidth]{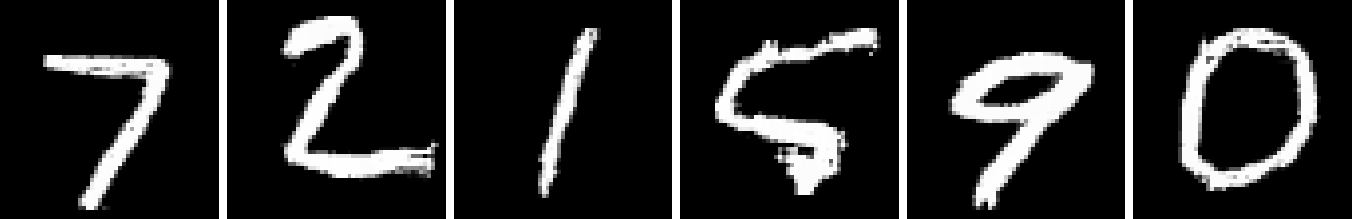}
    \caption{Reconstructions at $56 \times 56$}
    \label{fig:reconstructions-56x56}
    \vspace{0.1cm}
  \end{subfigure}
  \begin{subfigure}[b]{1\linewidth}
    \centering
    \includegraphics[width=1.0\linewidth]{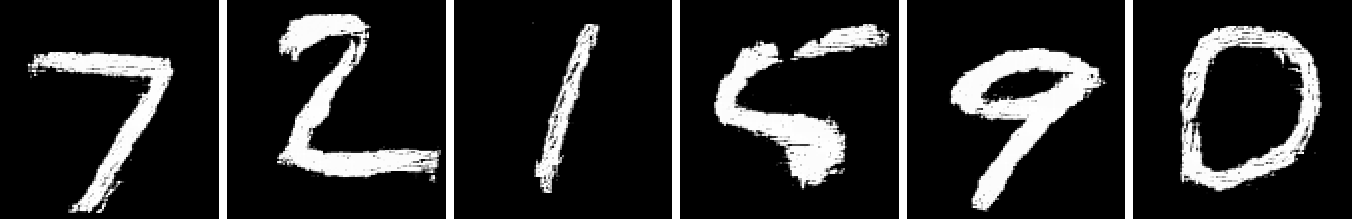}
    \caption{Reconstructions at $112 \times 112$}
    \label{fig:reconstructions-112x112}
    \vspace{0.1cm}
  \end{subfigure}
  \begin{subfigure}[b]{1\linewidth}
    \centering
    \includegraphics[width=1.0\linewidth]{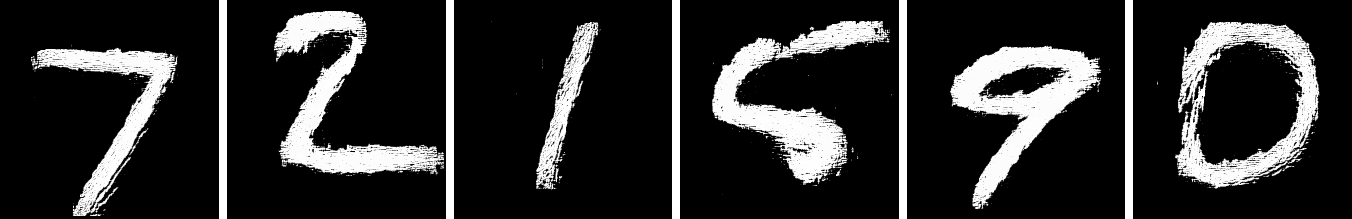}
    \caption{Reconstructions at $224 \times 224$}
    \label{fig:reconstructions-224x224}
  \end{subfigure}
  \caption{
    Reconstructions from a Spatial PixelCNN trained on $8 \times 8$ patches. 
    While VAE reconstructions are blurry (b), Spatial PixelCNN reconstructs the input quite convincingly with fine local details at the original $28 \times 28$ resolution (c).
    The model shows an impressive ability to produce coherent, plausible reconstructions at arbitrarily high resolutions; (d--f) show reconstructions up to resolution $224 \times 224$---an $8\times$ upscaling.
    Though note the increase in striations as the resolution increases.
    See Appendix~\ref{app:large-upscaling} for $20 \times$ and $50 \times$ reconstructions.
  }
  \label{fig:reconstructions}
\end{figure}

\begin{table*}[t]
\vskip 0.15in
\begin{center}
\begin{small}
\begin{sc}
\begin{tabular}{ccc|M{1.75cm}M{1.75cm}|M{1.75cm}M{1.75cm}}
\belowspace
& \multicolumn{5}{c}{PixelCNN++ without VAE} & \\
\hline
\abovespace\belowspace
& & & \multicolumn{2}{c|}{MS-SSIM} & \multicolumn{2}{c}{Confidence} \\
Patch Size & Coordinates & Bpd & 28x28 & 56x56 & 28x28 & 56x56 \\
\hline
\abovespace
\nextrow{row:28x28}$28 \times 28$\tast &         & \B 0.88 &    0.17 $\pm$ 0.26 &    0.66 $\pm$ 0.16 &    95.6 $\pm$ 1.2 &    70.7 $\pm$ 5.1 \\
\belowspace
$28 \times 28$ & $\surd$ &    0.89 & \B 0.17 $\pm$ 0.27 & \B 0.48 $\pm$ 0.29 & \B 97.5 $\pm$ 0.7 & \B 86.3 $\pm$ 3.3 \\
\hline
\abovespace
\nextrow{row:20x20}$20 \times 20$ &         &    1.50 &    0.09 $\pm$ 0.25 & \B 0.15 $\pm$ 0.19 &    89.6 $\pm$ 2.6 & \B 82.1 $\pm$ 3.6 \\
\belowspace
$20 \times 20$ & $\surd$ & \B 1.48 & \B 0.18 $\pm$ 0.26 &    0.27 $\pm$ 0.15 & \B 93.7 $\pm$ 1.7 &    81.5 $\pm$ 3.5 \\
\hline
\abovespace
\nextrow{row:16x16}$16 \times 16$ &         &    1.68 &    0.03 $\pm$ 0.24 &    0.02 $\pm$ 0.12 &    87.0 $\pm$ 2.9 &    79.9 $\pm$ 3.7 \\
\belowspace
$16 \times 16$ & $\surd$ & \B 1.65 & \B 0.14 $\pm$ 0.25 & \B 0.15 $\pm$ 0.18 & \B 91.1 $\pm$ 2.3 & \B 82.6 $\pm$ 3.7 \\
\hline
\abovespace
\nextrow{row:12x12}$12 \times 12$ &         &    1.73 &    0.02 $\pm$ 0.24 &    0.02 $\pm$ 0.13 & \B 86.4 $\pm$ 3.0 &    80.3 $\pm$ 3.8 \\
\belowspace
$12 \times 12$ & $\surd$ & \B 1.70 & \B 0.13 $\pm$ 0.24 & \B 0.20 $\pm$ 0.19 &    85.6 $\pm$ 3.5 & \B 83.3 $\pm$ 3.5 \\
\hline
\abovespace
\nextrow{row:4x4}$ 4 \times  4$ &         &    1.52 &    0.05 $\pm$ 0.21 &    0.05 $\pm$ 0.14 &    71.8 $\pm$ 6.0 &    76.3 $\pm$ 4.3 \\
\belowspace
$ 4 \times  4$ & $\surd$ & \B 1.42 & \B 0.14 $\pm$ 0.23 & \B 0.17 $\pm$ 0.19 & \B 79.1 $\pm$ 5.5 & \B 86.4 $\pm$ 2.9 \\
\hline
\end{tabular}
\caption{A comparison of PixelCNN++ trained with and without conditioning on a coordinate grid (as denoted by a $\surd$) for various patch sizes. MS-SSIM for the subset of $500$ MNIST test images is $0.16 \pm 0.27$, and the LeNet Confidence for the subset of $1000$ MNIST test images is $99.6 \pm 0.1$. \textsuperscript{*}PixelCNN++ baseline}
\label{tab:sample-analysis}
\begin{tabular}{ccc|M{1.75cm}M{1.75cm}|M{1.75cm}M{1.75cm}}
\rule{0pt}{5ex}%  EXTRA vertical height
\belowspace
& \multicolumn{5}{c}{PixelCNN++ with VAE} & \\
\hline
\abovespace\belowspace
& & & \multicolumn{2}{c|}{MS-SSIM} & \multicolumn{2}{c}{Confidence} \\
Patch Size & Coordinates & Bpd & 28x28 & 56x56 & 28x28 & 56x56 \\
\hline
\abovespace\belowspace
\nextrow{row:16x16-vae}$16 \times 16$      &         &    $\leq$1.95 &    0.03 $\pm$ 0.23 &    0.03 $\pm$ 0.11 &    87.6 $\pm$ 2.8 &    81.9 $\pm$ 3.5 \\
\belowspace
\nextrow{row:16x16-vae-coord}$16 \times 16$      & $\surd$ &    $\leq$1.87 &    0.19 $\pm$ 0.27 & \B 0.16 $\pm$ 0.21 &    93.4 $\pm$ 1.6 &    91.3 $\pm$ 2.3 \\
\belowspace
\nextrow{row:16x16-vae-coord-large}$16 \times 16$\tast & $\surd$ & \B $\leq$1.39 & \B 0.18 $\pm$ 0.27 &    0.15 $\pm$ 0.21 & \B 97.0 $\pm$ 0.8 &    93.2 $\pm$ 1.7 \\
\belowspace
\nextrow{row:8x8-vae-coord}$ 8 \times  8$      & $\surd$ &    $\leq$1.67 & \B 0.18 $\pm$ 0.27 &    0.13 $\pm$ 0.20 &    95.1 $\pm$ 1.3 & \B 94.1 $\pm$ 1.6 \\
\hline
\end{tabular}
\end{sc}
\end{small}
\end{center}
\vskip -0.1in
\caption{Analysis of PixelCNN++ combined with a VAE. We report the same metrics as in Table~\ref{tab:sample-analysis}. \textsuperscript{*}Uses the large network.}
\label{tab:vae-sample-analysis}
\end{table*}

Here we describe the details for the experiments conducted.
This includes model architecture, training hyperparameters, and the specific metrics used to evaluate the model.

\subsection{Model Details}

We use a modified version of PixelCNN++ \cite{salimans2017pixelcnn++}.
Our model similarly makes use of six blocks of ResNet \cite{he2016deep} layers.
Spatial PixelCNN is conditioned on both a latent vector and a regularized coordinate grid.
For each change in dimensions between blocks, the coordinate grid is resampled to the new layer's dimensions.
This resampling is performed as a simple linear interpolation.
Given a grid patch defined by $m \times m$ coordinates in the range $[(i_1, j_1), (i_m, j_m)]$, we linearly interpolate the range into $\frac{m}{2} \times \frac{m}{2}$ equal sized steps.
This regularization is key to preserving spatial coherence between ResNet blocks.

We additionally make use of ResNet blocks for the encoder and decoder of our VAE.
For each ResNet layer of the decoder, we condition on the latent vector $\latentZ$.

For most of our MNIST \cite{lecun1998gradient} experiments we use two ResNet layers for each block, with a convolution filter size of $25$.
We term this the \emph{small network}.
In order to determine the trade-off between the model's ability to compress data (as determined by having a smaller expected bits per dimension), versus its ability to effectively upscale images, we also conduct an experiment with a \emph{large network}.
This large network utilizes five ResNet layers for each block, with a convolution filter size of $140$.
The size of the latent code is fixed at $60$ for both network sizes.

\subsection{Training Details}

We make use of the Tensorflow \cite{tensorflow2015-whitepaper} framework for training and evaluating our models.
We train using the Adam \cite{kingma2014adam} optimizer, with a batch size of $128$, and an initial learning rate of of $0.001$ which is annealed using exponential decay at a rate of $0.999995$ every batch.
Additionally a dropout rate of $0.5$, along with L2 regularization of $0.0001$ is utilized.
We train each model until convergence, as determined by a lack of improvement over one hundred epochs in the model's bits per dimension.

At testing and image generation time we use the exponential moving average over the model parameters \cite{polyak1992acceleration}, calculated with a decay of $0.9995$ at each parameter update.
We report our negative log-likelihood values in bits per dimension.
The reported bits per dimension is calculated as the average bits per dimension over all possible patches for the images in the test set.

\subsection{Evaluation Details}

As previously noted \cite{theis2015note}, the negative log-likelihood may not be a great qualitative measure for evaluating generative models.
This can be seen clearly in our experiments, where bits per dimension does not accurately correlate with the ability of the model to generate coherent images (Fig.~\ref{fig:image-defects}). 

\textbf{MS-SSIM:} We first assess the diversity of the images generated by utilizing Multiscale Structural Similarity \cite{wang2003multiscale,odena2016conditional} (MS-SSIM).
To calculate the MS-SSIM for a given model, we begin by randomly sampling $500$ images from the model.
We produce an MS-SSIM score for all unique pairs of images.
We then report the mean and standard deviation of these scores.
The lower the MS-SSIM, the more diverse the images in the set.
The ideal MS-SSIM of a given model should closely resemble that of the underlying data, so we also report an MS-SSIM for $500$ images from the MNIST test set for comparison.

While a measure of diversity shows the model does not exhibit mode collapse, it is unable to address the subjective assessment of image quality.
Such an assessment should include the ability of the model to accurately reconstruct a given input, as well as ensuring generated samples reflect the underlying data distribution.
Thus we devise two additional measures.

\textbf{LeNet Accuracy:} As the combination of PixelCNN and VAE allows for conditioning on an interpretable latent representation, we can assess reconstruction accuracy for these models.
We take inspiration from the use of the Inception model \cite{odena2016conditional} for assessing accuracy.
As our model is trained on MNIST, we instead train a version of LeNet \cite{lecun1998gradient}, which has demonstrated strong classification ability for this dataset.
We reconstruct $1000$ images from the test set using our models, then measure the classification accuracy for these reconstructions.
For comparison, we also report the classification accuracy for the original $1000$ images from the test set.

\textbf{LeNet Confidence:} In order to assess random samples generated from the model, we devise a simple confidence score inspired by the Inception Score \cite{salimans2016improved}.
As part of the Inception Score measures diversity, we reformulate our metric to only measure the conditional probability of the label, given an image.
For a given image, we calculate the softmax of the LeNet logits.
The index of the largest value indicates the predicted class.
We take the largest value multiplied by one hundred as a confidence measure, indicating how confident the model is in the prediction.
We report the mean and standard deviation of this value for $1000$ random samples from each model.

Given that our model additionally shows strong ability to upscale images to higher resolutions, we also report results for generating $56 \times 56$ images.
For both classification accuracy and confidence, we first downsample images to $28 \times 28$ before running the LeNet classifier.
We note that individually these metrics are imperfect, especially when assessing downsampled images, but when taken in aggregate they yield a more holistic assessment.

\section{Results}

\begin{table}[t]
\vskip 0.15in
\begin{center}
\begin{small}
\begin{sc}
\begin{tabular}{ccM{1.75cm}M{1.75cm}}
\belowspace
& \multicolumn{2}{c}{PixelCNN++ with VAE} & \\
\hline
\abovespace\belowspace
& & \multicolumn{2}{c}{Accuracy} \\
Patch Size & Coordinates & 28x28 & 56x56 \\
\hline
\abovespace\belowspace
\nextrow{row:16x16-vae-reconst}$16 \times 16$      &         &    33.2  &    12.1  \\
\hline
\abovespace\belowspace
\nextrow{row:16x16-vae-coord-reconst}$16 \times 16$      & $\surd$ &    97.2  &    87.8  \\
\hline
\abovespace\belowspace
\nextrow{row:16x16-vae-large-reconst}$16 \times 16$\tast & $\surd$ & \B 98.2  &    91.4  \\
\hline
\abovespace\belowspace
\nextrow{row:8x8-vae-coord-reconst}$ 8 \times  8$      & $\surd$ &    96.2  & \B 95.3  \\
\hline
\end{tabular}
\end{sc}
\end{small}
\end{center}
\vskip -0.1in
\caption{A comparison of reconstruction accuracy of PixelCNN++ combined with VAE for various patch sizes (conditioning on a coordinate grid denoted by a $\surd$). LeNet Accuracy for the subset of $1000$ MNIST test images is $99.0$. \textsuperscript{*}Uses the large network.
}
\label{tab:reconstruction-analysis}
\end{table}

In this section we detail the results of our experiments.
As the model we propose has multiple components, we perform ablation experiments to verify the need for each.
When comparing MS-SSIM scores, we consider scores close to those computed for the actual MNIST digits to be better scores.
For LeNet Accuracy and Confidence, a higher score is considered better.

\subsection{Effects of Spatial Conditioning}

We first assess the importance of spatial conditioning on the generative ability of PixelCNN++ (Table~\ref{tab:sample-analysis}).
We train PixelCNN++ on various patch sizes, both with and without spatial conditioning.
We note that even when trained on full-sized images, conditioning on a grid of coordinates boosts both MS-SSIM and LeNet Confidence (Table~\ref{tab:sample-analysis} Row~\ref{row:28x28}).
This implies the addition of coordinates may help capture structure versus a baseline PixelCNN++.

As we sweep across patch sizes, we see the addition of coordinates keeps the MS-SSIM within range of the underlying dataset.
Though, we note as patch size decreases, we observe a decrease in confidence scores for the $28\times28$ samples, and associated coherence of the resultant images (Fig.~\ref{fig:image-defects-c}).

An important observation from the experiments is that the trained bits per dimension value is not a great representation of the quality of random samples generated from the models (Figs.~\ref{fig:image-defects-b} \& \ref{fig:image-defects-c}).
In the case of training on $4 \times 4$ patches (Table~\ref{tab:sample-analysis} Row~\ref{row:4x4}), the model achieves a similar ability to compress the data (as denoted by a low bits per dimension), as training on $20 \times 20$ patches (Table~\ref{tab:sample-analysis} Row~\ref{row:20x20}).
Though, comparing the LeNet Confidence of $28 \times 28$ generated images, the $20 \times 20$ patches clearly model the underlying dataset more accurately (training on $20 \times 20$ patches achieves a Confidence of $93.7 \pm 1.7$, while training on $4 \times 4$ patches only results in a score of $79.1 \pm 5.5$).

\subsection{Addition of a Latent Code}

We next consider the effects provided by the addition of a latent code.
Even without spatial conditioning, the model trained on $16 \times 16$ patches produces higher confidence scores with the addition of a VAE (Table~\ref{tab:sample-analysis} Row~\ref{row:16x16} \& Table~\ref{tab:vae-sample-analysis} Row~\ref{row:16x16-vae}).
Though, the model has trouble with scale (Fig.~\ref{fig:image-defects-d}).
This follows our intuition that spatial conditioning is key to capturing regularity across images.

\subsection{Images from Patches}

We now examine Spatial PixelCNN conditioned on both coordinates and a latent code.
While a direct comparison is difficult, we note that Spatial PixelCNN produces samples qualitatively similar to PixelCNN++ (Fig.~\ref{fig:random-samples}).
Additionally Spatial PixelCNN trained on $8 \times 8$ patches receives a similar Confidence score to the PixelCNN++ baseline when generating $28 \times 28$ images (Table~\ref{tab:sample-analysis} Row~\ref{row:28x28} \& Table~\ref{tab:vae-sample-analysis} Row~\ref{row:8x8-vae-coord}).

An interesting result can be observed when comparing the small network trained on $8 \times 8$ patches versus the small network trained on $16 \times 16$ patches.
Given an equivalent network size, training on smaller patches produces superior results (Table~\ref{tab:vae-sample-analysis} Rows~\ref{row:16x16-vae-coord} \& \ref{row:8x8-vae-coord}).
Only with the addition of more parameters, through the use of the large network, does training on $16 \times 16$ patches produce a higher confidence and accuracy (Table~\ref{tab:vae-sample-analysis} Rows~\ref{row:16x16-vae-coord-large} \& \ref{row:8x8-vae-coord}) when generating $28 \times 28$ images.

\subsection{Upscaling Ability}

Finally we assess the ability for the network to generate high resolution images while only training on a low resolution dataset.
Looking at the generated $56 \times 56$ images, the small network trained on $8 \times 8$ patches works exceedingly well for upscaling (Fig.~\ref{fig:random-samples-small}).
It even tends to retain structural detail better than the large network trained on $16 \times 16$ patches (Fig.~\ref{fig:random-samples-large}).
Not only does it surpass the accuracy and confidence scores of the large network (Table~\ref{tab:vae-sample-analysis} Rows~\ref{row:16x16-vae-coord-large} \& \ref{row:8x8-vae-coord} and Table~\ref{tab:reconstruction-analysis} Rows \ref{row:16x16-vae-large-reconst} \& \ref{row:8x8-vae-coord-reconst}), it only loses a small amount of accuracy and confidence, despite upscaling $2\times$.
It even shows remarkable ability for large upscaling factors (Figs.~\ref{fig:concept}, \ref{fig:reconstructions-112x112}, \& \ref{fig:reconstructions-224x224}).
See Appendix~\ref{app:large-upscaling} for examples of $20 \times$ and $50 \times$ upscaling.

\section{Discussion \& Conclusions}

We demonstrated that the addition of coordinates to PixelCNN++ has a clear positive effect on its generative ability.
This can be seen even when trained on full-sized images, rather than on patches.
We also showed that with the combination of a VAE, Spatial PixelCNN trained on patches is able to not only generate convincing images at the same resolution as the underlying dataset, but also high resolution images, despite only training on a low resolution dataset.

We believe that our approach should allow for training on mixed-sized images by feeding a resampled version of the image to the VAE.
This may potentially allow for less preprocessing of images used for training.
We leave this open for future research.

\section{Acknowledgments}

We thank Jeff Clune for his valuable discussions and support. 
We also thank the University of Wyoming Advanced Research Computing Center (ARCC) for their assistance and computing resources which enabled us to perform experiments and analyses.
Additionally, we appreciate OpenAI making their code for PixelCNN++ available for reference online.

\bibliography{references}

\begin{thebibliography}{41}
\providecommand{\natexlab}[1]{#1}
\providecommand{\url}[1]{\texttt{#1}}
\expandafter\ifx\csname urlstyle\endcsname\relax
  \providecommand{\doi}[1]{doi: #1}\else
  \providecommand{\doi}{doi: \begingroup \urlstyle{rm}\Url}\fi

\bibitem[Abadi et~al.(2015)Abadi, Agarwal, Barham, Brevdo, Chen, Citro,
  Corrado, Davis, Dean, Devin, Ghemawat, Goodfellow, Harp, Irving, Isard, Jia,
  Jozefowicz, Kaiser, Kudlur, Levenberg, Man\'{e}, Monga, Moore, Murray, Olah,
  Schuster, Shlens, Steiner, Sutskever, Talwar, Tucker, Vanhoucke, Vasudevan,
  Vi\'{e}gas, Vinyals, Warden, Wattenberg, Wicke, Yu, and
  Zheng]{tensorflow2015-whitepaper}
Abadi, Mart\'{\i}n, Agarwal, Ashish, Barham, Paul, Brevdo, Eugene, Chen,
  Zhifeng, Citro, Craig, Corrado, Greg~S., Davis, Andy, Dean, Jeffrey, Devin,
  Matthieu, Ghemawat, Sanjay, Goodfellow, Ian, Harp, Andrew, Irving, Geoffrey,
  Isard, Michael, Jia, Yangqing, Jozefowicz, Rafal, Kaiser, Lukasz, Kudlur,
  Manjunath, Levenberg, Josh, Man\'{e}, Dan, Monga, Rajat, Moore, Sherry,
  Murray, Derek, Olah, Chris, Schuster, Mike, Shlens, Jonathon, Steiner,
  Benoit, Sutskever, Ilya, Talwar, Kunal, Tucker, Paul, Vanhoucke, Vincent,
  Vasudevan, Vijay, Vi\'{e}gas, Fernanda, Vinyals, Oriol, Warden, Pete,
  Wattenberg, Martin, Wicke, Martin, Yu, Yuan, and Zheng, Xiaoqiang.
\newblock {TensorFlow}: Large-scale machine learning on heterogeneous systems,
  2015.
\newblock URL \url{https://www.tensorflow.org/}.
\newblock Software available from tensorflow.org.

\bibitem[Arjovsky et~al.(2017)Arjovsky, Chintala, and Bottou]{Arjovsky:2017aa}
Arjovsky, Martin, Chintala, Soumith, and Bottou, L{\'e}on.
\newblock Wasserstein gan.
\newblock 01 2017.
\newblock URL \url{https://arxiv.org/abs/1701.07875}.

\bibitem[Chen et~al.(2016)Chen, Kingma, Salimans, Duan, Dhariwal, Schulman,
  Sutskever, and Abbeel]{chen2016variational}
Chen, Xi, Kingma, Diederik~P, Salimans, Tim, Duan, Yan, Dhariwal, Prafulla,
  Schulman, John, Sutskever, Ilya, and Abbeel, Pieter.
\newblock Variational lossy autoencoder.
\newblock \emph{arXiv preprint arXiv:1611.02731}, 2016.

\bibitem[Clune \& Lipson(2011)Clune and Lipson]{clune2011evolving}
Clune, Jeff and Lipson, Hod.
\newblock Evolving 3d objects with a generative encoding inspired by
  developmental biology.
\newblock \emph{ACM SIGEVOlution}, 5\penalty0 (4):\penalty0 2--12, 2011.

\bibitem[Denton et~al.(2015)Denton, Chintala, Szlam, and Fergus]{Denton:2015aa}
Denton, Emily, Chintala, Soumith, Szlam, Arthur, and Fergus, Rob.
\newblock Deep generative image models using a laplacian pyramid of adversarial
  networks.
\newblock 06 2015.
\newblock URL \url{https://arxiv.org/abs/1506.05751}.

\bibitem[Goodfellow et~al.(2014)Goodfellow, Pouget-Abadie, Mirza, Xu,
  Warde-Farley, Ozair, Courville, and Bengio]{goodfellow2014generative}
Goodfellow, Ian, Pouget-Abadie, Jean, Mirza, Mehdi, Xu, Bing, Warde-Farley,
  David, Ozair, Sherjil, Courville, Aaron, and Bengio, Yoshua.
\newblock Generative adversarial nets.
\newblock In \emph{Advances in neural information processing systems}, pp.\
  2672--2680, 2014.

\bibitem[Graves(2013)]{graves2013generating}
Graves, Alex.
\newblock Generating sequences with recurrent neural networks.
\newblock \emph{arXiv preprint arXiv:1308.0850}, 2013.

\bibitem[Gulrajani et~al.(2016)Gulrajani, Kumar, Ahmed, Taiga, Visin, Vazquez,
  and Courville]{gulrajani2016pixelvae}
Gulrajani, Ishaan, Kumar, Kundan, Ahmed, Faruk, Taiga, Adrien~Ali, Visin,
  Francesco, Vazquez, David, and Courville, Aaron.
\newblock Pixelvae: A latent variable model for natural images.
\newblock \emph{arXiv preprint arXiv:1611.05013}, 2016.

\bibitem[Ha(2016)]{ha2017latent}
Ha, David.
\newblock Generating large images from latent vectors.
\newblock
  \url{http://blog.otoro.net/2016/04/01/generating-large-images-from-latent-vectors/},
  2016.

\bibitem[He et~al.(2016)He, Zhang, Ren, and Sun]{he2016deep}
He, Kaiming, Zhang, Xiangyu, Ren, Shaoqing, and Sun, Jian.
\newblock Deep residual learning for image recognition.
\newblock In \emph{Proceedings of the IEEE conference on computer vision and
  pattern recognition}, pp.\  770--778, 2016.

\bibitem[Kalchbrenner et~al.(2016)Kalchbrenner, Espeholt, Simonyan, Oord,
  Graves, and Kavukcuoglu]{kalchbrenner2016neural}
Kalchbrenner, Nal, Espeholt, Lasse, Simonyan, Karen, Oord, Aaron van~den,
  Graves, Alex, and Kavukcuoglu, Koray.
\newblock Neural machine translation in linear time.
\newblock \emph{arXiv preprint arXiv:1610.10099}, 2016.

\bibitem[Kalchbrenner et~al.(2017)Kalchbrenner, van~den Oord, Simonyan,
  Danihelka, Vinyals, Graves, and Kavukcuoglu]{pmlr-v70-kalchbrenner17a}
Kalchbrenner, Nal, van~den Oord, A{\"a}ron, Simonyan, Karen, Danihelka, Ivo,
  Vinyals, Oriol, Graves, Alex, and Kavukcuoglu, Koray.
\newblock Video pixel networks.
\newblock In Precup, Doina and Teh, Yee~Whye (eds.), \emph{Proceedings of the
  34th International Conference on Machine Learning}, volume~70 of
  \emph{Proceedings of Machine Learning Research}, pp.\  1771--1779,
  International Convention Centre, Sydney, Australia, 06--11 Aug 2017. PMLR.
\newblock URL \url{http://proceedings.mlr.press/v70/kalchbrenner17a.html}.

\bibitem[Karras et~al.(2017)Karras, Aila, Laine, and
  Lehtinen]{karras2017progressive}
Karras, Tero, Aila, Timo, Laine, Samuli, and Lehtinen, Jaakko.
\newblock Progressive growing of gans for improved quality, stability, and
  variation.
\newblock \emph{arXiv preprint arXiv:1710.10196}, 2017.

\bibitem[Kim et~al.(2016)Kim, Jernite, Sontag, and Rush]{kim2016character}
Kim, Yoon, Jernite, Yacine, Sontag, David, and Rush, Alexander~M.
\newblock Character-aware neural language models.
\newblock In \emph{AAAI}, pp.\  2741--2749, 2016.

\bibitem[Kingma \& Ba(2014)Kingma and Ba]{kingma2014adam}
Kingma, Diederik and Ba, Jimmy.
\newblock Adam: A method for stochastic optimization.
\newblock \emph{arXiv preprint arXiv:1412.6980}, 2014.

\bibitem[Kingma \& Welling(2013)Kingma and Welling]{Kingma:2013aa}
Kingma, Diederik~P and Welling, Max.
\newblock Auto-encoding variational bayes.
\newblock 12 2013.
\newblock URL \url{https://arxiv.org/abs/1312.6114}.

\bibitem[Krause et~al.(2016)Krause, Lu, Murray, and Renals]{Krause:2016aa}
Krause, Ben, Lu, Liang, Murray, Iain, and Renals, Steve.
\newblock Multiplicative lstm for sequence modelling.
\newblock 09 2016.
\newblock URL \url{https://arxiv.org/abs/1609.07959}.

\bibitem[Larochelle \& Murray(2011)Larochelle and Murray]{larochelle2011neural}
Larochelle, Hugo and Murray, Iain.
\newblock The neural autoregressive distribution estimator.
\newblock In \emph{Proceedings of the Fourteenth International Conference on
  Artificial Intelligence and Statistics}, pp.\  29--37, 2011.

\bibitem[LeCun et~al.(1998)LeCun, Bottou, Bengio, and
  Haffner]{lecun1998gradient}
LeCun, Yann, Bottou, L{\'e}on, Bengio, Yoshua, and Haffner, Patrick.
\newblock Gradient-based learning applied to document recognition.
\newblock \emph{Proceedings of the IEEE}, 86\penalty0 (11):\penalty0
  2278--2324, 1998.

\bibitem[Ledig et~al.(2016)Ledig, Theis, Husz{\'a}r, Caballero, Cunningham,
  Acosta, Aitken, Tejani, Totz, Wang, et~al.]{ledig2016photo}
Ledig, Christian, Theis, Lucas, Husz{\'a}r, Ferenc, Caballero, Jose,
  Cunningham, Andrew, Acosta, Alejandro, Aitken, Andrew, Tejani, Alykhan, Totz,
  Johannes, Wang, Zehan, et~al.
\newblock Photo-realistic single image super-resolution using a generative
  adversarial network.
\newblock \emph{arXiv preprint arXiv:1609.04802}, 2016.

\bibitem[Nguyen et~al.(2015)Nguyen, Yosinski, and Clune]{nguyen2015deep}
Nguyen, Anh, Yosinski, Jason, and Clune, Jeff.
\newblock Deep neural networks are easily fooled: High confidence predictions
  for unrecognizable images.
\newblock In \emph{Proceedings of the IEEE Conference on Computer Vision and
  Pattern Recognition}, pp.\  427--436, 2015.

\bibitem[Nguyen et~al.(2017)Nguyen, Yosinski, Bengio, Dosovitskiy, and
  Clune]{nguyen2017plug}
Nguyen, Anh, Yosinski, Jason, Bengio, Yoshua, Dosovitskiy, Alexey, and Clune,
  Jeff.
\newblock Plug \& play generative networks: Conditional iterative generation of
  images in latent space.
\newblock \emph{Proceedings of the IEEE Conference on Computer Vision and
  Pattern Recognition}, 2017.

\bibitem[Nowozin et~al.(2016)Nowozin, Cseke, and Tomioka]{nowozin2016f}
Nowozin, Sebastian, Cseke, Botond, and Tomioka, Ryota.
\newblock f-gan: Training generative neural samplers using variational
  divergence minimization.
\newblock In \emph{Advances in Neural Information Processing Systems}, pp.\
  271--279, 2016.

\bibitem[Odena et~al.(2016)Odena, Olah, and Shlens]{odena2016conditional}
Odena, Augustus, Olah, Christopher, and Shlens, Jonathon.
\newblock Conditional image synthesis with auxiliary classifier gans.
\newblock \emph{arXiv preprint arXiv:1610.09585}, 2016.

\bibitem[Oord et~al.(2016{\natexlab{a}})Oord, Dieleman, Zen, Simonyan, Vinyals,
  Graves, Kalchbrenner, Senior, and Kavukcuoglu]{oord2016wavenet}
Oord, Aaron van~den, Dieleman, Sander, Zen, Heiga, Simonyan, Karen, Vinyals,
  Oriol, Graves, Alex, Kalchbrenner, Nal, Senior, Andrew, and Kavukcuoglu,
  Koray.
\newblock Wavenet: A generative model for raw audio.
\newblock \emph{arXiv preprint arXiv:1609.03499}, 2016{\natexlab{a}}.

\bibitem[Oord et~al.(2016{\natexlab{b}})Oord, Kalchbrenner, and
  Kavukcuoglu]{oord2016pixel}
Oord, Aaron van~den, Kalchbrenner, Nal, and Kavukcuoglu, Koray.
\newblock Pixel recurrent neural networks.
\newblock \emph{arXiv preprint arXiv:1601.06759}, 2016{\natexlab{b}}.

\bibitem[Oord et~al.(2017)Oord, Vinyals, and Kavukcuoglu]{oord2017neural}
Oord, Aaron van~den, Vinyals, Oriol, and Kavukcuoglu, Koray.
\newblock Neural discrete representation learning.
\newblock \emph{arXiv preprint arXiv:1711.00937}, 2017.

\bibitem[Polyak \& Juditsky(1992)Polyak and Juditsky]{polyak1992acceleration}
Polyak, Boris~T and Juditsky, Anatoli~B.
\newblock Acceleration of stochastic approximation by averaging.
\newblock \emph{SIAM Journal on Control and Optimization}, 30\penalty0
  (4):\penalty0 838--855, 1992.

\bibitem[Rezende et~al.(2014)Rezende, Mohamed, and
  Wierstra]{pmlr-v32-rezende14}
Rezende, Danilo~Jimenez, Mohamed, Shakir, and Wierstra, Daan.
\newblock Stochastic backpropagation and approximate inference in deep
  generative models.
\newblock In Xing, Eric~P. and Jebara, Tony (eds.), \emph{Proceedings of the
  31st International Conference on Machine Learning}, volume~32 of
  \emph{Proceedings of Machine Learning Research}, pp.\  1278--1286, Bejing,
  China, 22--24 Jun 2014. PMLR.
\newblock URL \url{http://proceedings.mlr.press/v32/rezende14.html}.

\bibitem[Salimans et~al.(2016)Salimans, Goodfellow, Zaremba, Cheung, Radford,
  and Chen]{salimans2016improved}
Salimans, Tim, Goodfellow, Ian, Zaremba, Wojciech, Cheung, Vicki, Radford,
  Alec, and Chen, Xi.
\newblock Improved techniques for training gans.
\newblock In \emph{Advances in Neural Information Processing Systems}, pp.\
  2234--2242, 2016.

\bibitem[Salimans et~al.(2017)Salimans, Karpathy, Chen, and
  Kingma]{salimans2017pixelcnn++}
Salimans, Tim, Karpathy, Andrej, Chen, Xi, and Kingma, Diederik~P.
\newblock Pixelcnn++: Improving the pixelcnn with discretized logistic mixture
  likelihood and other modifications.
\newblock \emph{arXiv preprint arXiv:1701.05517}, 2017.

\bibitem[Secretan et~al.(2008)Secretan, Beato, D~Ambrosio, Rodriguez, Campbell,
  and Stanley]{secretan2008picbreeder}
Secretan, Jimmy, Beato, Nicholas, D~Ambrosio, David~B, Rodriguez, Adelein,
  Campbell, Adam, and Stanley, Kenneth~O.
\newblock Picbreeder: evolving pictures collaboratively online.
\newblock In \emph{Proceedings of the SIGCHI Conference on Human Factors in
  Computing Systems}, pp.\  1759--1768. ACM, 2008.

\bibitem[Stanley(2007)]{stanley2007compositional}
Stanley, Kenneth~O.
\newblock Compositional pattern producing networks: A novel abstraction of
  development.
\newblock \emph{Genetic programming and evolvable machines}, 8\penalty0
  (2):\penalty0 131--162, 2007.

\bibitem[Stanley et~al.(2009)Stanley, D'Ambrosio, and
  Gauci]{stanley2009hypercube}
Stanley, Kenneth~O, D'Ambrosio, David~B, and Gauci, Jason.
\newblock A hypercube-based encoding for evolving large-scale neural networks.
\newblock \emph{Artificial life}, 15\penalty0 (2):\penalty0 185--212, 2009.

\bibitem[Theis \& Bethge(2015)Theis and Bethge]{theis2015generative}
Theis, Lucas and Bethge, Matthias.
\newblock Generative image modeling using spatial lstms.
\newblock In \emph{Advances in Neural Information Processing Systems}, pp.\
  1927--1935, 2015.

\bibitem[Theis et~al.(2015)Theis, Oord, and Bethge]{theis2015note}
Theis, Lucas, Oord, A{\"a}ron van~den, and Bethge, Matthias.
\newblock A note on the evaluation of generative models.
\newblock \emph{arXiv preprint arXiv:1511.01844}, 2015.

\bibitem[Tyka(2017)]{tyka2017semantic}
Tyka, Mike.
\newblock Superresolution with semantic guide.
\newblock
  \url{https://mtyka.github.io/machine/learning/2017/08/09/highres-gan-faces-followup.html},
  2017.

\bibitem[van~den Oord et~al.(2016)van~den Oord, Kalchbrenner, Espeholt,
  Vinyals, Graves, et~al.]{van2016conditional}
van~den Oord, Aaron, Kalchbrenner, Nal, Espeholt, Lasse, Vinyals, Oriol,
  Graves, Alex, et~al.
\newblock Conditional image generation with pixelcnn decoders.
\newblock In \emph{Advances in Neural Information Processing Systems}, pp.\
  4790--4798, 2016.

\bibitem[Wang et~al.(2003)Wang, Simoncelli, and Bovik]{wang2003multiscale}
Wang, Zhou, Simoncelli, Eero~P, and Bovik, Alan~C.
\newblock Multiscale structural similarity for image quality assessment.
\newblock In \emph{Signals, Systems and Computers, 2004. Conference Record of
  the Thirty-Seventh Asilomar Conference on}, volume~2, pp.\  1398--1402. IEEE,
  2003.

\bibitem[Werbos(1990)]{werbos1990backpropagation}
Werbos, Paul~J.
\newblock Backpropagation through time: what it does and how to do it.
\newblock \emph{Proceedings of the IEEE}, 78\penalty0 (10):\penalty0
  1550--1560, 1990.

\bibitem[Zhang et~al.(2016)Zhang, Xu, Li, Zhang, Huang, Wang, and
  Metaxas]{zhang2016stackgan}
Zhang, Han, Xu, Tao, Li, Hongsheng, Zhang, Shaoting, Huang, Xiaolei, Wang,
  Xiaogang, and Metaxas, Dimitris.
\newblock Stackgan: Text to photo-realistic image synthesis with stacked
  generative adversarial networks.
\newblock \emph{arXiv preprint arXiv:1612.03242}, 2016.

\end{thebibliography}
\bibliographystyle{icml2017}

\onecolumn
\appendix
\appendixpage
\section{Large Upscaling Factors}
\label{app:large-upscaling}

\subsection{Upscaling Comparisons}
\label{app:large-upscaling-comparison}

\begin{figure*}[ht]
  \centering
  \begin{subfigure}[b]{1\linewidth}
    \centering
    \includegraphics[width=1.0\linewidth]{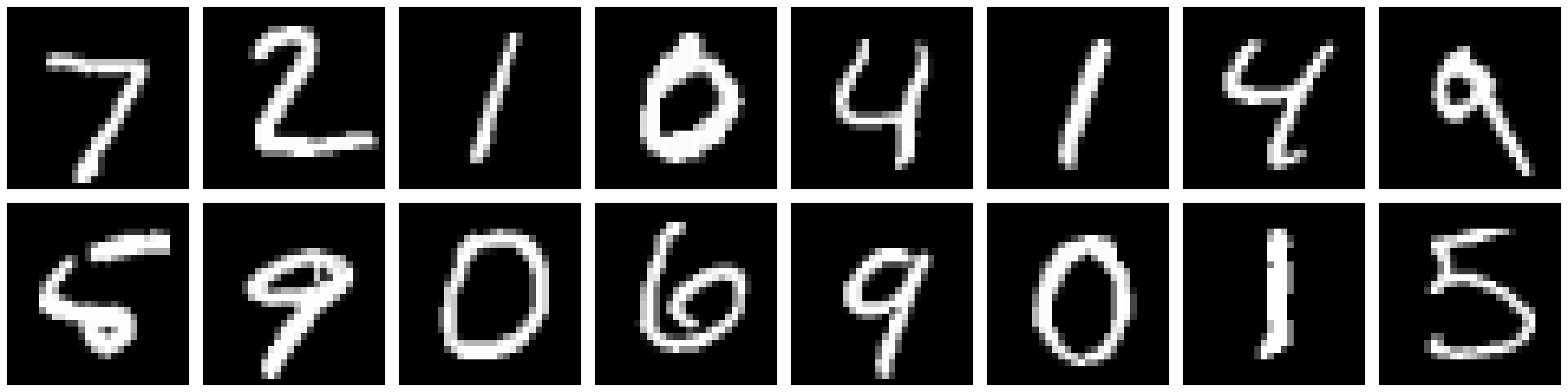}
    \caption{MNIST test images}
    \label{fig:large-originals-28x28}
    \vspace{0.1cm}
  \end{subfigure}
  \begin{subfigure}[b]{1\linewidth}
    \centering
    \includegraphics[width=1.0\linewidth]{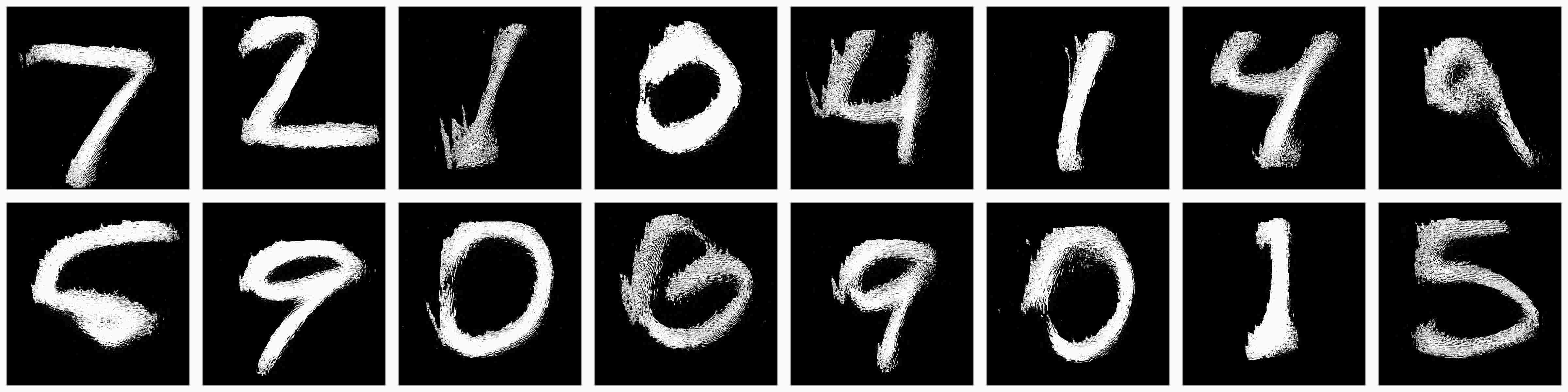}
    \caption{Reconstructions at $560 \times 560$ (a $20\times$ upscaling)}
    \label{fig:large-reconstructions-560x560}
    \vspace{0.1cm}
  \end{subfigure}
  \begin{subfigure}[b]{1\linewidth}
    \centering
    \includegraphics[width=1.0\linewidth]{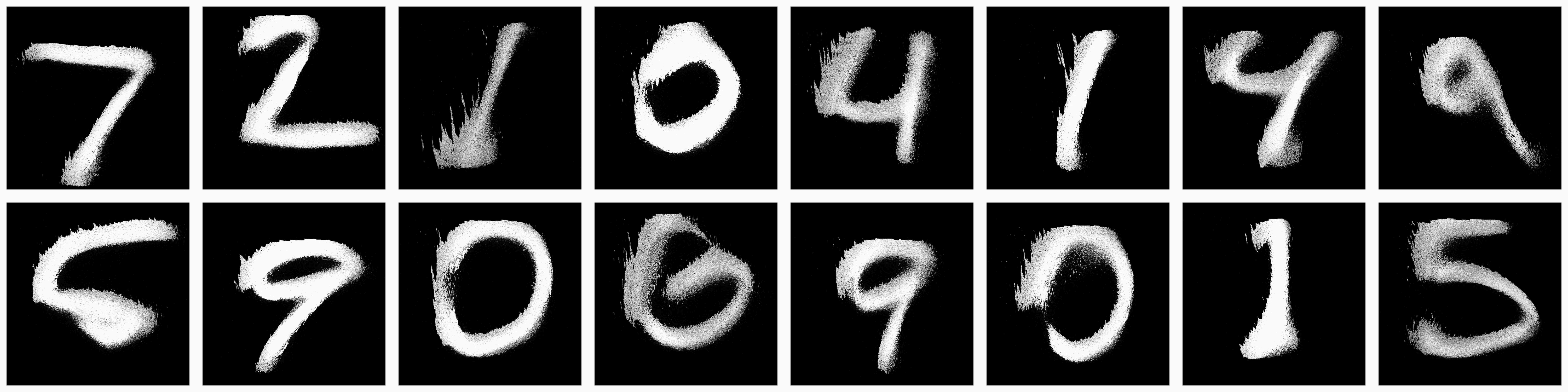}
    \caption{Reconstructions at $1400 \times 1400$ (a $50\times$ upscaling)}
    \label{fig:large-reconstructions-1400x1400}
    \vspace{0.1cm}
  \end{subfigure}
  \caption{
    Reconstructions from Spatial PixelCNN (small network trained on $8 \times 8$ patches) at very large resolutions.
    The images are clearly identifable as MNIST digits.
  }
  \label{fig:large-reconstructions}
\end{figure*}

\subsection{A Single High-Resolution Example}
\label{app:large-upscaling-example}

\begin{figure*}[ht]
  \centering
  \includegraphics[width=1.0\linewidth]{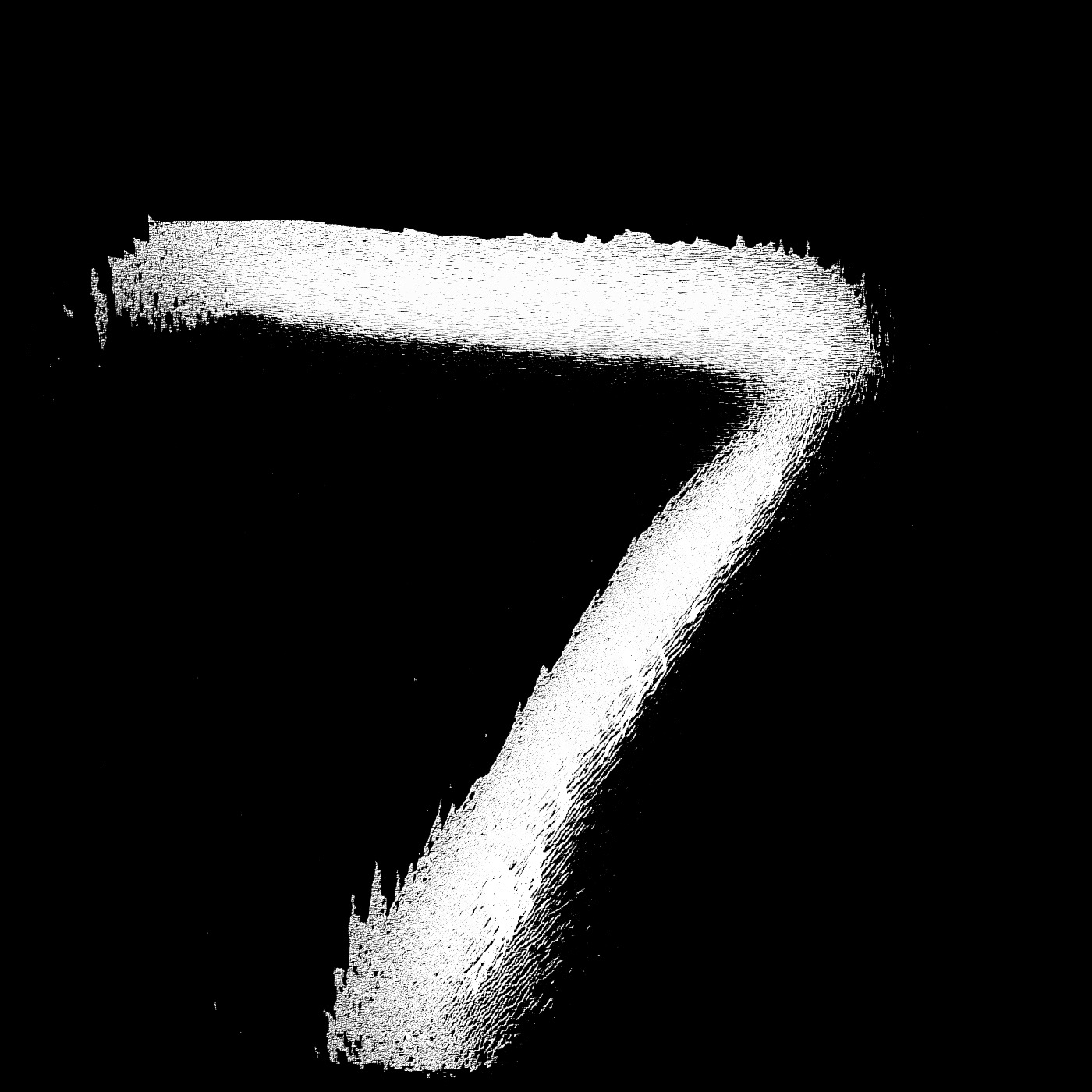}
  \caption{Reconstruction at $1400 \times 1400$ (a $50 \times$ upscaling) from Spatial PixelCNN (small network trained on $8 \times 8$ patches).}
  \label{fig:single-large-reconstruction}
\end{figure*}

\end{document}